\documentclass[10pt,a4paper,conference]{IEEEtran}
\usepackage[caption=false]{subfig}
\usepackage{graphicx}
\usepackage{svg}
\usepackage{multirow}
\usepackage{xcolor}
\usepackage{amsfonts}   
\usepackage{amsmath}
\usepackage{comment}
\usepackage{pifont}

\newcommand\blfootnote[1]{%
  \begingroup
  \renewcommand\thefootnote{}\footnote{#1}%
  \addtocounter{footnote}{-1}%
  \endgroup
}

\ifCLASSINFOpdf
\else
\fi
\begin{document}
%
\title{DAG-Net: Double Attentive Graph Neural Network for Trajectory Forecasting}

\author{\IEEEauthorblockN{Alessio Monti*, Alessia Bertugli*, Simone Calderara and Rita Cucchiara}
\IEEEauthorblockA{AImageLab, University of Modena and Reggio Emilia, Modena, Italy\\
Email: \{alessio.monti, alessia.bertugli, simone.calderara, rita.cucchiara\}@unimore.it}}


%


\maketitle

\begin{abstract}

Understanding human motion behaviour is a critical task for several possible applications like self-driving cars or social robots, and in general for all those settings where an autonomous agent has to navigate inside a human-centric environment. This is non-trivial because human motion is inherently multi-modal: given a history of human motion paths, there are many plausible ways by which people could move in the future. Additionally, people activities are often driven by goals, e.g. reaching particular locations or interacting with the environment. We address the aforementioned aspects by proposing a new recurrent generative model that considers both single agents' future goals and interactions between different agents. The model exploits a double attention-based graph neural network to collect information about the mutual influences among different agents and to integrate it with data about agents' possible future objectives. Our proposal is general enough to be applied to different scenarios: the model achieves state-of-the-art results in both urban environments and also in sports applications. \blfootnote{*Equal contribution.}
\end{abstract}

%
\IEEEpeerreviewmaketitle

\section{Introduction}
Trajectory prediction has become an essential component for several applications: self-driving cars and social robots may leverage useful insights about human motion to preventively forecast pedestrian actions and avoid collisions \cite{forecasting_interactive_dynamics, multi_agents_fusion, glmp-realtime, goal_robots}, while surveillance systems can benefit from knowing how crowds will move to better monitor huddled environments \cite{surveillance_1, abnormal_crowd}. There is also a high interest outside the smart cities context, like for example in team sports, where such predictions could give important insights for tactical analysis. However, designing a model to help to predict agents' trajectories is as desirable as difficult: a series of demanding challenges has to be faced.

First of all, the task results particularly tough because human motion is inherently \textit{multi-modal}: when moving, people may follow several plausible trajectories, as there is a rich distribution of potential human behaviours. This means that, given a particular set of past observations, there is no unique correct future since several behaviours could be equally appropriate \cite{sgan, social-bigat}. In an urban environment, a pedestrian could choose to reach his/her destination following several plausible paths: cut straight to the objective crossing the road, or maybe take a longer walk following physical clues like sidewalks and pedestrian crossings. Similarly, in a basketball match, when an attacking player is running towards an opponent, several modes of behaviour develop: the player may choose to avoid the defender by passing the ball to a teammate, or perhaps achieve the same result by dribbling the adversary directly. Furthermore, each of these possible modes may exhibit a high variance in return: agents can modify step after step some of their features, like speed.

\begin{figure}[!t]
\resizebox{\columnwidth}{!}{
    \centering
    \includegraphics[width=\textwidth]{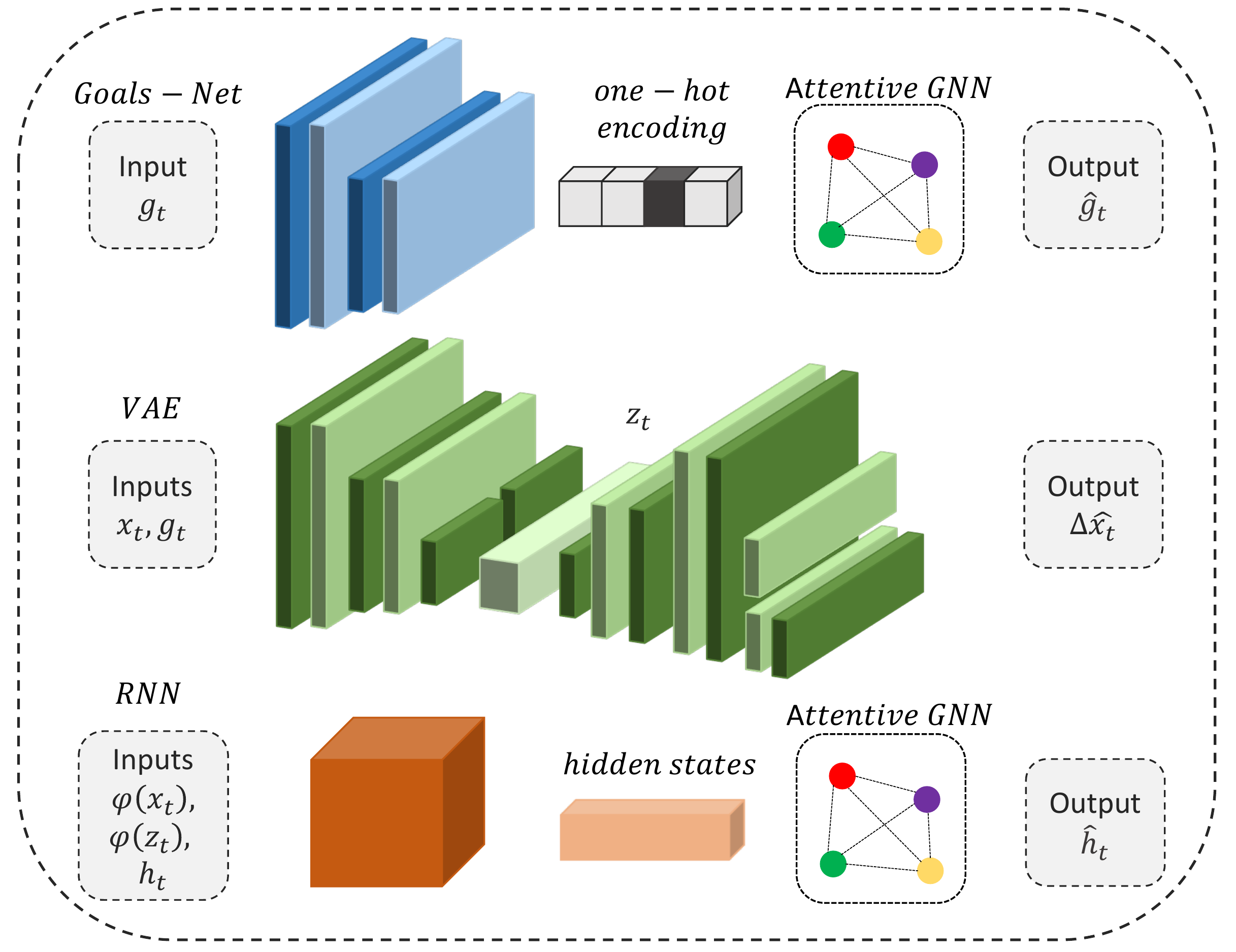}}
    \caption{Scheme of DAG-Net architecture. It is composed of a Goal-Net that learns to predict agents' future goals; a VAE to generate displacements at every time-step; a RNN to consider the temporal nature of the sequence.}
    \label{fig:model}
\end{figure}

Another challenge, especially when agents move in crowded scenarios, is represented by \textit{social interactions}. Interactions heavily impact on future trajectories \cite{sgan, slstm}: since people plan their paths reading each other future possible behaviours, each person's motion is influenced by the subjects around them. Interactions do not have to be necessarily as explicit as walking in groups or talking to each other: a moving pedestrian can interact with other people just by implicitly considering the positions of the surrounding agents to avoid future collisions. In team sports, interactions take on an even more important role: an attacker could put in place a certain set of movements just because the rest of his team has a specific disposition, as well as the whole defending team could in turn react and put in place a predefined tactic only because some opponents are arranged in a certain way. The varied nature of interactions and their heavy impact on agents' behaviour leads to develop sophisticated methods to accurately interpret and integrate such information into the prediction method.

Even the \emph{future knowledge} about interacting agents positions can be a relevant feature that affects the development of each path. Taking into account this aspect during the prediction can improve the accuracy of the model.

To address these challenges we propose DAG-Net, a double attentive graph neural network for trajectory forecasting. The network backbone is a recurrent version of the Variational Autoencoder (VAE)~\cite{vae}: time-step after time-step, the autoencoder is used to generate the future position in terms of the displacement from the current location. The modules of our recurrent autoencoder are conditioned on subjects' objectives so that the model can accordingly produce likely future positions. The backbone is integrated with a double Graph Neural Network (GNN)-based mechanism: the first GNN defines the future objectives of each agent in a structured way, distilling each goal with proximity knowledge; the second GNN models agents' interactions, filtering the hidden states of the recurrent network through neighbourhood information. Both the GNNs use a self-attention mechanism to assign different weights to each edge of the graph. The entire model is depicted in Fig. \ref{fig:model}.

\section{Related Works}

Literature extensively demonstrated that trajectory prediction cannot leave aside modelling the interactions between different agents. 
Early works in trajectory prediction took advantage of hand-crafted features and energy potential parameters \cite{socialforce, discrete_choice, continuum_dynamics, gaussian_process}. In particular, Helbing and Molnar \cite{socialforce} modelled an interaction between two agents as a social force: the idea is to use attractive forces to guide agents toward their destination while employing repulsive forces to encourage collision avoidance. Other noteworthy examples are the Discrete Choice framework by Antonini et al. \cite{discrete_choice}, Continuum Dynamics by Treuille et al.
\cite{continuum_dynamics}, and Gaussian Processes by Wang et al. \cite{gaussian_process}. Although these methods exhibit robust performance, they share a common weakness: the drawback is represented by the very same hand-crafted features, which fail to generalise properly and struggle in complex scenarios, limiting the results in terms of prediction accuracy.
More modern approaches \cite{slstm, soft+, desire} rely on recurrent neural networks (RNNs) or one of their more efficient variants (LSTMs or GRUs) to learn these features directly from data: such architectures are specifically designed to exploit the temporal dependencies that characterise time-series; hence they are particularly suited for predicting trajectories, where every position is strictly correlated with the previous. To model all the agents inside a particular scene, the na\"ive use of such models foresees the employing of a single RNN with shared parameters. However, without further solutions, such a network would predict single agents trajectories independently.
To share information across different subjects, several mechanisms have been proposed.

In this sense, one of the most important contributions is Social Pooling by Alahi et al. \cite{slstm}: the mechanism merges agents' recurrent hidden states inside a \textit{social tensor}. To build the social tensor the model employs a \textit{grid-based pooling}: given an agent $i$ and its neighbourhood $\mathcal{N}_{i}$, all the hidden states of agent $i$'s neighbours are pooled together inside the social tensor, which is then fed to the recurrent cell as one of the inputs. The main limitation of this approach is the neighbourhood itself: such a local solution fails to capture the global context, as it does not allow the model to consider the interactions between all the possible agents inside the scene in a computationally efficient manner.

Social GAN by Gupta et al. \cite{sgan} introduces, inside the Generator, a new Pooling Module that instead combines information coming from all the possible agents present in the scene. For every agent $i$, all the hidden states $\mathbf{h}_{t}^{j}$ of the other agents are processed by an MLP and max-pooled together element-wise into the tensor $P_{i}$.

Social Ways by Amirian et al. \cite{socialways} takes up and customise this solution: instead of a simple max-pooling, the influence of the other agents on the generic agent $i$ is evaluated by applying an attention weighting procedure. The model builds a so-called \textit{interaction feature vector}: the $i^{th}$ vector is created by combining new social features that come from predefined geometric properties. Given two agents $i$ and $j$, the vector is built by stacking information like the Euclidean distance between the agents or the bearing angle from agent $j$ to agent $i$. A simple scalar product between the $i^{th}$ interaction vector and the hidden $\mathbf{h}^{j}$ brings to the attention coefficient $a^{ij}$.

Another relevant work is STGAT by Huang et al. \cite{stgat}. Instead of employing custom pooling modules, the authors exploited the recent progress in Graph Neural Networks (GNNs). To share information across different pedestrians, STGAT treats each agent as a node of a graph. In this manner, the model can employ a GNN \cite{gat} as its sharing mechanism, allowing to aggregate information from neighbours by performing self-attention \cite{attention_is_all_you_need} on graph nodes.

A different approach comes instead from Zhan et al. \cite{weeksup}: rather than sharing the hidden states, the authors tried to induce coordination by working on agents' future intentions. The method is described as \textit{weakly supervised}, since it requires a preliminary extraction from ground-truth trajectories of some low-dimensional features (called \textit{macro-intents}) that will hopefully provide a tractable way to capture the coordination between agents. By conditioning the generation of the future intent on the hidden state of a recurrent cell that watches the whole set of agents, the produced macro-intents will somehow be also influenced by what other agents are willing to do. This allows the model to keep in consideration the coordination between the different subjects.

DAG-Net is inspired by \cite{weeksup} since it employs a similar idea of intents. However, our objectives are generated and used in a different way. We treat goals as structured components by exploiting them as graph nodes and we accordingly condition the generation process on the resulting interrelated future objectives. Furthermore, we use an attentive module to associate importance weights to different interactive nodes.
\section{Method}

\subsection{Problem definition}
The position of a generic agent $i$ at time $t$ is represented by $\mathbf{x}_{t}^{i}=(x^{i}, y^{i})_{t}$, where $(x^{i}, y^{i})_{t}$ are the coordinates that localise the agent in the scene at the given time-step; agent $i$'s trajectory can be therefore defined as a series $X_{i} = \{\mathbf{x}_{1}^{i},...,\mathbf{x}_{T}^{i}\}$ of consecutive positions. Every trajectory $X_{i}$ is split into past and future: given a certain number $T_{obs}$ of past positions, the goal is to predict in most accurate way the next $T_{pred}$ future positions. All the coordinates are taken with respect to a real-world reference system, thus are expressed in meters or feet and do not refer to single pixels.

Before feeding the data to the model, every trajectory is transformed into a series of relative positions: every absolute position $\mathbf{x}_{t}^{i}$ is transformed in a couple of displacements $(\Delta x_{t}^{i}, \Delta y_{t}^{i})$ that express the movement along the two axes with respect to the previous absolute position $\mathbf{x}_{t-1}^{i}$. Coming back to the original coordinates is always possible: given the initial absolute position $\mathbf{x}_{1}^{i}$, the other absolute coordinates can be obtained with a cumulative sum along the remaining time-steps. To simplify the notation, we use $\mathbf{x}_t$ instead of $\Delta x_{t}$ to denote a displacement at time-step $t$.

\subsection{Recurrent VAE}
The network backbone is realised with a recurrent version of the Variational Autoencoder \cite{deep_gen_models}. For non-sequential data, Variational Autoencoders have already been shown to be effective in recovering and modelling complex multi-modal distributions over the data space: for this purpose, a VAE introduces a set of latent random variables $\mathbf{z}$ within the latent space $\mathcal{Z}$, specifically designed to capture the characterising variations in the observed input variables $\mathbf{x}$.

In order to model and generate sequences that can be both highly variable and highly structured (e.g. trajectories), Chung et al. \cite{vrnn} extended this approach by proposing a recurrent version of the VAE, namely the Variational Recurrent Neural Network (VRNN). The network retains the necessary flexibility to model highly non-linear dynamics, but also explicitly models the dependencies between latent variables across subsequent time-steps. The VRNN can be described as a combination of the VAE and the RNN architectures: more specifically, the VRNN contains a variational autoencoder for every time-step of the input sequence, whose prior distribution over $\mathbf{z}$ is conditioned on the hidden state $\mathbf{h}_{t}$ of a common RNN. The combination with a recurrent cell helps the variational autoencoder architecture to keep into consideration the temporal structure of the input data. Generally, the model is employed in a completely generative setting: after training, the model is used to generate brand new sequences that however resemble the original dataset examples. Since we want to accurately reproduce agents' future path, we instead employ the model in a predictive setting: after a burn-in period of $T_{obs}$ observation time-steps, the model is used to generate $T_{pred}$ new future positions that have to be as close as possible to the ground-truth positions we want to forecast.

The model can be decomposed in four sub-components: the encoder, the decoder, the prior, and the RNN, that are implemented like neural networks as $\varphi^{enc}$, $\varphi^{dec}$, $\varphi^{prior}$ and $\varphi^{rnn}$. 

The first stage is represented by the encoder: the encoder receives raw data (in our case, single $\mathbf{x}_{t}$ displacements at a given time-step), embeds them in a fixed-length feature vector, incorporates the vector with the last hidden state $\mathbf{h}_{t-1}$ of the recurrent cell, and obtains the representation of their combination in the latent space $\mathcal{Z}$. Thus, the approximate posterior will not only be a function of $\mathbf{x}_{t}$ as in the VAE but also a function of $\mathbf{h}_{t-1}$:
\begin{align}
\label{eq:vrnn_encoder}
    \pmb{\mu}_{\mathbf{z},t}, \pmb{\sigma}_{\mathbf{z},t} &= \varphi^{enc}\left(\varphi^{\mathbf{x}}(\mathbf{x}_{t}), \mathbf{h}_{t-1}\right), \\
    \label{eq:vrnn_infernce}
    q_{\phi}\left(\mathbf{z}_{t} | \mathbf{x}_{\leq t}, \mathbf{z}_{<t}\right) &=\mathcal{N}\left(\mathbf{z}_{t} | \boldsymbol{\mu}_{\mathbf{z},t},\left(\boldsymbol{\sigma}_{\mathbf{z},t }\right)^{2}\right),
\end{align}
where $\varphi^{\mathbf{x}}$ is a neural network that extracts the features from the input and $\phi$ are the approximation function parameters.

The decoder network takes the latent variable $\mathbf{z}_{t}$, embeds it in a fixed-size vector, incorporates the embedding with the last hidden state $\mathbf{h}_{t-1}$ of the recurrent cell, and provides a reconstruction $\mathbf{\hat{x}}_t$ of  $\mathbf{x}_t$:
\begin{align}
    \label{eq:vrnn_decoder}
    \pmb{\mu}_{\mathbf{\hat{x}},t} \pmb{\sigma}_{\mathbf{\hat{x}},t} &= \varphi^{dec}\left(\varphi^{\mathbf{z}}(\mathbf{z}_{t}), \mathbf{h}_{t-1}\right), \\
    \label{eq:vrnn_generation}
    p_{\theta}\left(\mathbf{x}_{t} | \mathbf{x}_{<t}, \mathbf{z}_{\leq t}\right) &= \mathcal{N}\left(\mathbf{x}_{t} | \boldsymbol{\mu}_{\mathrm{\mathbf{\hat{x}},t}},\left(\boldsymbol{\sigma}_{\mathbf{\hat{x}}, t}\right)^{2}\right),
\end{align}
where $\varphi^{\mathbf{z}}$ is a feature extracting neural network and $\theta$ are the approximation function parameters.
In the end, the objective of the decoder is to output a sample $\mathbf{\hat{x}}_{t}$ that resembles as much as possible the original input $\mathbf{x}_{t}$.

The prior network is instead able to reach the latent space $\mathcal{Z}$ starting only from the last hidden state $\mathbf{h}_{t-1}$ of the recurrent cell. It is computed as follow:
\begin{align}
  \label{eq:norm_prior_vrnn}
    \pmb{\mu}_{\mathbf{0},t}, \pmb{\sigma}_{\mathbf{0},t} &= \varphi^{prior}\left(\mathbf{h}_{t-1}\right), \\
    \label{eq:prior_vrnn}
    p_{\theta}\left(\mathbf{z}_{t} | \mathbf{x}_{<t}, \mathbf{z}_{<t}\right) &= \mathcal{N}\left(\mathbf{z}_{t} | \boldsymbol{\mu}_{\mathbf{0},t},\left(\boldsymbol{\sigma}_{\mathbf{0},t}\right)^{2}\right).
\end{align}

The prior is essential to \textit{generate} new data: to take a step forward and not just reconstruct the input (e.g. when we want to predict the next future displacement), the encoding network is detached. Here the prior network ensures that we are still able to reach the latent space $\mathcal{Z}$, even without the encoder: furthermore, since the produced latent variables resemble the ones returned by the encoder, passing them to the decoder allows to obtain likely (yet new) examples, as if they were actual inputs coming from the dataset.

Finally, the RNN updates its hidden state by taking into account both the input $\mathbf{x}_{t}$ and the latent variable $\mathbf{z}_{t}$: this encourages the explicit modelling of the temporal dependencies across subsequent time-steps.
\begin{align}
    \label{eq:vrnn_recurrence}
    \mathbf{h}_{t} &= \varphi^{rnn}\left(\mathbf{x}_{t}, \mathbf{z}_{t}, \mathbf{h}_{t-1}\right)
\end{align}

The entire model is trained by maximising the sequential \textit{evidence lower-bound} (ELBO):

\begin{equation}
\label{eq:elbo_vrnn}
\begin{split}
    &\mathbb{E}_{q_{\phi}(\mathbf{z}_{\leq T} \mid \mathbf{x}_{\leq T})} 
    \left [
        \sum^{T} \log p_{\theta}(\mathbf{x}_{t} \mid \mathbf{z}_{\leq t}, \mathbf{x}_{<t}) \right. \\
            &
            \left. - D_{KL} \left ( 
                q_{\phi}(\mathbf{z}_{t} \mid \mathbf{x}_{\leq t}, \mathbf{z}_{< t}) \mid \mid p_{\theta}(\mathbf{z}_{t} \mid \mathbf{x}_{< t}, \mathbf{z}_{< t})
            \right ) 
    \vphantom{\sum^{T}} \right ].
\end{split}
\end{equation}

\bigbreak
The loss can be interpreted as the variational autoencoder ELBO summed over each time-step $t$ of the input sequence.

\subsection{Conditioning VAE to agents' goals}
\begin{figure}[!t]%
   \centering
   \vspace{3pt}
    \subfloat[]{\includegraphics[scale=0.223]{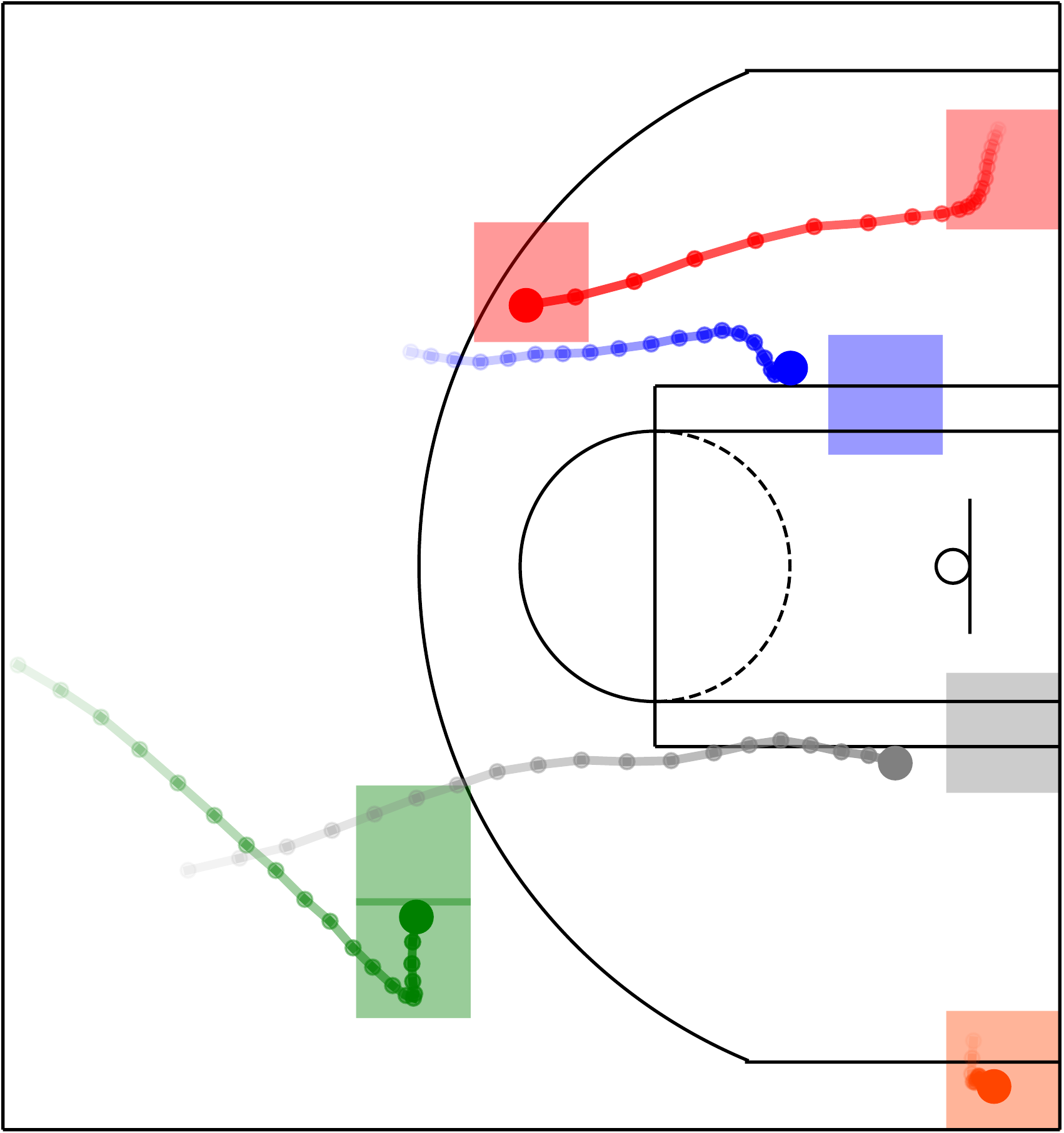}\label{fig:goals_plot}}\qquad
    \subfloat[]{\includegraphics[scale=0.223]{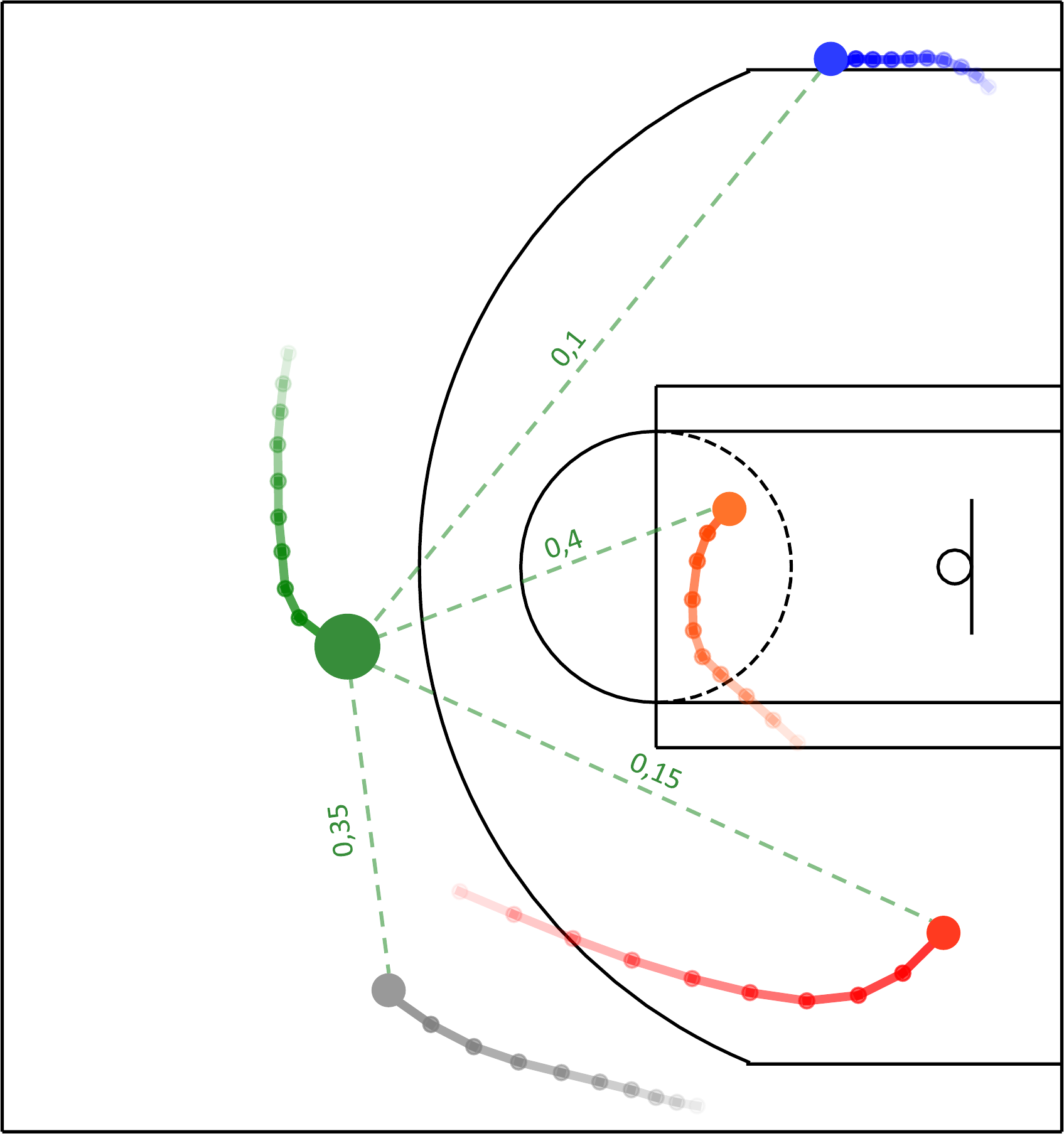}\label{fig:similarities_plot}}\qquad
    \caption{In (a) we can observe how goals deeply influence past and future trajectories, guiding agents to specific portions of the court. In (b) we can observe the similarities between the green player and his teammates: these values will directly influence the recombination of both goals and hidden states at the green node.}
    \label{fig:goals_and_similaritiesa}
\end{figure}

Inspired by \cite{cvae,cvae2,weeksup,conditionalflow, where_will_they_go,Jiachen_IROS19}, we provide an additional input to our backbone in order to condition the displacements generation process to agents' future objectives.
We choose to describe agents' future goals in terms of spatial information (Fig. \ref{fig:goals_plot}). To make our model as invariant as possible with respect to the different characteristics of the environment, we divided the top-down view of the scene in a grid of macro-areas: each cell can potentially represent the future objective of a single agent. 
Agent $i$'s goal at time $t$, $\mathbf{g^i_t}$, is then represented by a one-hot encoding of the grid, where the cell in which the agent will land in the future is filled with a 1. 

To obtain ground-truth objectives, a sliding window approach has been used: a window of size $w$ slides through the original absolute trajectory and captures a goal every $w$ time-steps. This information is used to condition the \emph{prior}, the \emph{encoder} and the \emph{decoder} networks, as shown in Eq. \eqref{eq:norm_pri}, \eqref{eq:norm_enc} and \eqref{eq:norm_dec} where we drop the superscripts to refer to the behaviour of a general agent.

\begin{align}
    \label{eq:norm_pri}
    \pmb{\mu}_{\mathbf{0},t}, \pmb{\sigma}_{\mathbf{0},t} &= \varphi^{prior}\left(\mathbf{h}_{t-1},  \mathbf{g}_{t}\right), \\
    \label{eq:norm_enc} 
    \pmb{\mu}_{\mathbf{z},t}, \pmb{\sigma}_{\mathbf{z},t} &= \varphi^{enc}\left(\varphi^{\mathbf{x}}(\mathbf{x}_{t}), \mathbf{h}_{t-1}, \mathbf{g}_{t}\right), \\
    \label{eq:norm_dec}
    \pmb{\mu}_{\mathbf{\hat{x}},t} \pmb{\sigma}_{\mathbf{\hat{x}},t} &= \varphi^{dec}\left(\varphi^{\mathbf{z}}(\mathbf{z}_{t}), \mathbf{h}_{t-1}, \mathbf{g}_{t}\right).
\end{align}

To produce likely goals during the inference phase, we employ a further network. This network is again conditioned on the hidden state $\mathbf{h}_{t-1}$ of the recurrent cell, and takes as additional inputs the last predicted objective for the agent and the concatenation $\mathbf{d}_{t-1}$ of the absolute positions of the other agents in the scene (i.e. their disposition).

\begin{equation}
    \label{eq:goals_net}
    \mathbf{g}'_{t} = \varphi^{goal}(\mathbf{g}'_{t-1}, \mathbf{d}_{t-1}, \mathbf{h}_{t-1})
\end{equation}

\begin{equation}
\label{eq:elbo_vrnn_goals}
\begin{split}
    &\mathbb{E}_{q_{\phi}(\mathbf{z}_{\leq T} \mid \mathbf{x}_{\leq T})} 
    \left [
        \sum_{t=1}^{T}\sum_{k=1}^{K} \log p_{\theta}(\mathbf{x}_{t} \mid \mathbf{z}_{\leq t}, \mathbf{x}_{<t}) - \mathbf{g}^{k}_{t} log(\mathbf{g}'^{k}_{t})\right. \\
            &
            \left. - D_{KL} \left ( 
                q_{\phi}(\mathbf{z}_{t} \mid \mathbf{x}_{\leq t}, \mathbf{z}_{< t}) \mid \mid p_{\theta}(\mathbf{z}_{t} \mid \mathbf{x}_{< t}, \mathbf{z}_{< t})
            \right ) 
    \vphantom{\sum^{T}} \right ] 
\end{split}
\end{equation}

The additional loss term is computed as a Cross-Entropy between the ground-truth goal $\mathbf{g}_{t}$ and the predicted one $\mathbf{g'}_{t}$, where $K$ is total the number of cells inside their one-hot encoding.

\subsection{Double Attentive Graph Neural Networks}
DAG-Net leverages two graph attentive networks to model two different kinds of interactions: the interactions between agents and the relationships between future goals. In structured motion environments, where agents' behaviours are moved not only by single intentions but also by social rules and/or common goals, it is important to condition the prediction to both mutual interactions and neighbours objectives. DAG-Net jointly employs past data and future intentions to improve forecasting in such contexts.

\subsubsection{Goals relationships}
As seen in Eq. \eqref{eq:norm_pri}, \eqref{eq:norm_enc} and \eqref{eq:norm_dec}, agents' future objectives are used to condition the backbone: nevertheless, without further solutions, a single predicted goal $\mathbf{g}'_{t}$ focuses only on the corresponding agent objective. To effectively capture the coordination between the different subjects in the scene, DAG-Net shares goals information among agents relying on group interactions. To model the structure of future interactions, an attentive GNN \cite{gat} is employed. At every time-step, the network takes as an input node the one-hot encoding of each agent's predicted goal $\mathbf{g}'_{t}$, and produces a new distilled goal $\mathbf{\Tilde{g}}_t$ built on proximity notions. After the concatenation of the distilled goal with the original one, the final refined goal is obtained through a linear projection:

\begin{equation}
    \label{eq:gg}
    \hat{\mathbf{g}}_{t} = W \left(\mathbf{g}'_{t} \mathbin\Vert \mathbf{\Tilde{g}}_{t}\right)
\end{equation}
where the parameter matrix $W \in \mathcal{R}^{dxd} $ (with $d$ the number of goals grid cells) is learnt in an end-to-end fashion during training. The new produced goal $\hat{\mathbf{g}}_{t}$ will then take the place of $\mathbf{g}'_{t}$ inside the ELBO loss presented in Eq. \eqref{eq:elbo_vrnn_goals}.

\subsubsection{Agents' interactions}
To model the interactions between the different agents in the scene, our model uses again a graph-based approach. Each agent is connected to the others as a node of a graph where edges weights are defined by a self-attention mechanism. A distance-based adjacency matrix is used in addition to the attentive module to consider proximity information (Fig. \ref{fig:similarities_plot}). At every time-step, the hidden state $\mathbf{h}_t$ of each agent is fed to the GNN: the network outputs a new distilled hidden state $\mathbf{\Tilde{h}}_t$ that takes into account the history of the neighbours. After the concatenation of the distilled hidden state with the original one, a linear layer is used to achieve the final output:

\begin{equation}
    \label{eq:hg}
    \hat{\mathbf{h}}_t = H \left(\mathbf{h}_t \mathbin\Vert \mathbf{\Tilde{h}}_t\right)
\end{equation}
where the parameter matrix $H$ is learnt during the training phase. The new refined hidden state $\hat{\mathbf{h}}_t$ will then be used in the next time-step as the current hidden state of the agent.

\subsection{Training protocol}

During training, we let the network see the entire $T = T_{obs}+T_{pred}$ time-steps from ground-truth sequences. The solution gives the model the opportunity to collect important features also from the latest time-steps of the sequence: this is useful in urban contexts and results particularly effective in sports, where we have long trajectories that usually start as linear but seldom continue in the same way, often bending and turning back upon themselves.

During validation and testing, we instead divide the trajectories into an \textit{observation} and a \textit{prediction} split: specifically, the network burns in for $T_{obs}$ time-steps observing the first portion of the ground-truth trajectory, then it's let predicting the remaining $T_{pred}$ time-steps.
\section{Experiments}
\subsection{Implementation Details}
The VRNN recurrent cell is a GRU with 1 recurrent layer and a hidden state dimension of 64; the dimension of the latent variable is set to 32. For each graph, we then employ two attentive GNN layers: the first layer reduces the input to lower-dimensional hidden space, the second layer returns instead to the original input space. Each GNN layer uses 4 attention heads. The entire model has been optimised with Adam optimiser. To cope with the differences between urban and sport settings, we employ different sets of hyper-parameters.

For the urban setting, we use a learning rate of $10^{-4}$ and a batch size of 16; the Cross-Entropy contribution is weighted with a factor of $10^{-2}$. The hidden state dimension between the two graph layers is set to 4. The model has been trained for 500 epochs.

For the sports setting, we use a learning rate of $10^{-3}$ and a batch-size of 64; the Cross-Entropy contribution is weighted with a factor of $10^{-2}$. The hidden state dimension between the two graph layers is set to 8. The model has been trained for 300 epochs.

\subsection{Metrics}
Similar to prior works in literature, the model is evaluated with respect to two error metrics on prediction results, \textit{Average Displacement Error} (ADE) and \textit{Final Displacement Error} (FDE).

The Average Displacement Error represents the average Euclidean distance, over the entire predicted sequence, between the ground-truth positions and the predicted ones:

\begin{equation}
    \label{eq:ade}
    ADE = \frac{
                \sum_{i \in \mathcal{P}} 
                \sum_{t=0}^{T_{pred}} 
                \sqrt{((\hat{x}_{t}^{i}, \hat{y}_{t}^{i}) - (x_{t}^{i}, y_{t}^{i}))^{2}}
            } 
            {
                \mid \mathcal{P} \mid \cdot \thinspace \thinspace T_{pred}
            }
\end{equation}

where $\mathcal{P}$ is the set of pedestrians considered, $\mid \mathcal{P} \mid$ is its cardinality, $(\hat{x}_{t}^{i}, \hat{y}_{t}^{i})$ are the predicted absolute coordinates at time $t$, and $(x_{t}^{i}, y_{t}^{i})$ are the ground-truth absolute coordinates at time $t$.

The Final Displacement Error follows the same logic but focuses only on the last time-step, evaluating the Euclidean distance between the final ground-truth position and the predicted one.

\begin{equation}
    \label{eq:fde}
    FDE = \frac{
                \sum_{i \in \mathcal{P}} 
                \sqrt{((\hat{x}_{T_{pred}}^{i}, \hat{y}_{T_{pred}}^{i}) - (x_{T_{pred}}^{i}, y_{T_{pred}}^{i}))^{2}}
            } 
            {
                \mid \mathcal{P} \mid
            }
\end{equation}

\subsection{Datasets}

\subsubsection{Stanford Drone Dataset}
The Stanford Drone Dataset \cite{sdd} is composed of a series of top-down videos recorded by a hovering drone in 8 different college campus scenes. This large scale dataset collects complex and crowded scenarios with various types of interacting targets: apart from classic pedestrians, we can also find bikes, skateboarders, cars, buses, and other vehicles, therefore the navigation inside such environments results particularly tough. We use the TrajNet benchmark version of the dataset \cite{trajnet2018}: trajectories are composed of a series of consecutive positions expressed as $(x,y)$ world coordinates and recorded at 2.5FPS. Due to the lack of annotation in the test set, we split the training set into three sub-sets for the training, test and validation phases.

\subsubsection{STATS SportVU NBA Dataset}
The dataset comes from the player tracking data provided by STATS SportVU \cite{stats_sportvu}. The dataset contains tracking positions from the 2016 NBA regular season on a span of over 1200 different games: the data are recorded with a series of cameras that surround the court and give back a bird-eye view of players' positions. The games are split into offensive plays, hence every sequence starts when the ball crosses the middle of the court. A play ends when one the following conditions is met: a shot is made (missed or scored), the ball exceeds the court bounds, the ball is intercepted by the defending team, or the shot clock runs out. Each play composes of 50 time-steps sampled at 5Hz, where each time-step contains the positions (expressed as $(x,y,z)$ world coordinates) for all the 10 players on the court (5 attackers, 5 defenders) plus the ball. All the data have been subsequently normalised and shifted to have zero-centred sequences to the middle of court and plays that always develop towards the right basket.

\subsection{Quantitative Results}
For the basketball setting, we evaluated separately offence and defence: since agents are placed in an explicit competitive setting, their nature is intrinsically different, both from the goals and the trajectories points of view. The attackers call the shots trying to score, while the defenders usually react to their moves: training the network simultaneously on both teams would distract the final results.

\begin{table}[!t]
    \renewcommand{\arraystretch}{1.3}
    \centering
    \caption{Basketball SportVU results}
     \label{sportvu_results}
    \begin{tabular}{c|c|cc}
        \hline
        \textbf{Team} & \textbf{Model} & \textbf{ADE} & \textbf{FDE}\\
        \hline \hline
                \multirow{4}{*}{ATK}&STGAT \cite{stgat} & 9.94 & 15.80\\
                & Social-Ways \cite{socialways} & 9.91 & 15.19\\
                & Weak-Supervision \cite{weeksup} & 9.47 & 16.98\\
                & DAG-Net (Our) & \textbf{8.98} & \textbf{14.08}\\
        \hline
                \multirow{4}{*}{DEF}&STGAT \cite{stgat} & 7.26 & 11.28\\
                & Social-Ways \cite{socialways} & 7.31 & 10.21\\
                & Weak-Supervision \cite{weeksup} & 7.05 & 10.56\\
                & DAG-Net (Our) & \textbf{6.87} & \textbf{9.76}\\
        \hline
    \end{tabular}
\end{table}

\begin{table}[!t]
    \renewcommand{\arraystretch}{1.3}
    \centering
    \caption{Long-term evaluations}
    \label{ours_vs_cvae_long_term}
    \begin{tabular}{c|c|ccc}
        \hline
         \multirow{2}{*}{\textbf{Model}} & \multirow{2}{*}{\textbf{Team}} & \textbf{20-10 Split} & \textbf{20-20 Split} &\textbf{20-30 Split} \\
         && \textbf{ADE} &  \textbf{ADE} &  \textbf{ADE} \\
        \hline \hline
                C-VAE \cite{where_will_they_go} & ATK & 3.95 & 5.80 & 7.08 \\
                DAG-Net (Our) & ATK & \textbf{2.09} & \textbf{4.58} & \textbf{6.66} \\
        \hline
                C-VAE \cite{where_will_they_go} & DEF & 3.01 & 4.10 & \textbf{4.98} \\
                DAG-Net (Our) & DEF & \textbf{2.05} & \textbf{4.07} & 5.01 \\
        \hline
    \end{tabular}
\end{table}

\begin{table}[!t]
    \renewcommand{\arraystretch}{1.3}
    \centering
    \caption{Stanford Drone Dataset results}
    \label{sdd_results}
    \begin{tabular}{c|cc}
        \hline
        \textbf{Model} & \textbf{ADE} & \textbf{FDE}\\
        \hline \hline
             STGAT \cite{stgat} & 0.58 & 1.11\\
             Social-Ways \cite{socialways} & 0.62 & 1.16\\      
             DAG-Net (Our) & \textbf{0.53} & \textbf{1.04}\\
        \hline
    \end{tabular}
\end{table}

The values reported in Table \ref{sportvu_results} support our choice, with defence trajectories being clearly easier than attacking ones. All the metrics are expressed in feet and refer to a prediction horizon of 40 time-steps, with 10 initial steps of observation. 

The model achieves impressive results compared to state-of-the-art methods. STGAT shows quite strong performances, being able to weight all the possible contributions from the agents in the scene, but lacks additional information about what could happen in the long-term. With its attention-based pooling, Social-Ways returns very similar results: the combination of its geometric social features gives useful clues on such complicated trajectories and interactions as the basketball ones. Even though its different approach, Weak Supervision gain promising results too: the exploitation of future intentions allow agents to plan correctly their future path. However, by jointly considering agents' interactions and future goals, our model is more able to capture the nature of real paths and to reach smaller errors with respect to all these competitive methods.

\begin{figure}[!t]
  \centering
  \vspace{4pt}
  \begin{minipage}[b]{0.24\textwidth}
   \includegraphics[width=\textwidth]{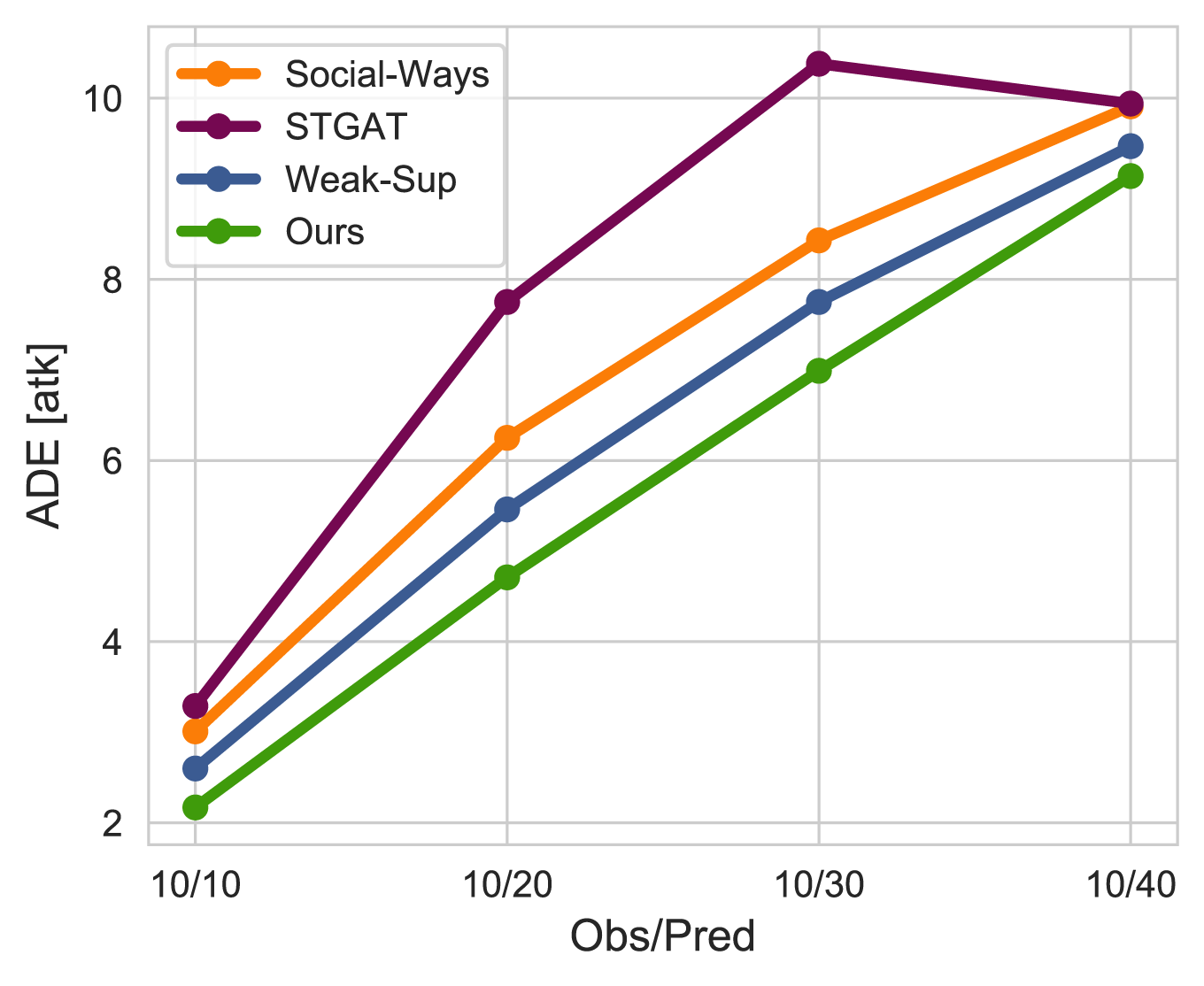}
  \end{minipage}
  \begin{minipage}[b]{0.24\textwidth}
    \includegraphics[width=\textwidth]{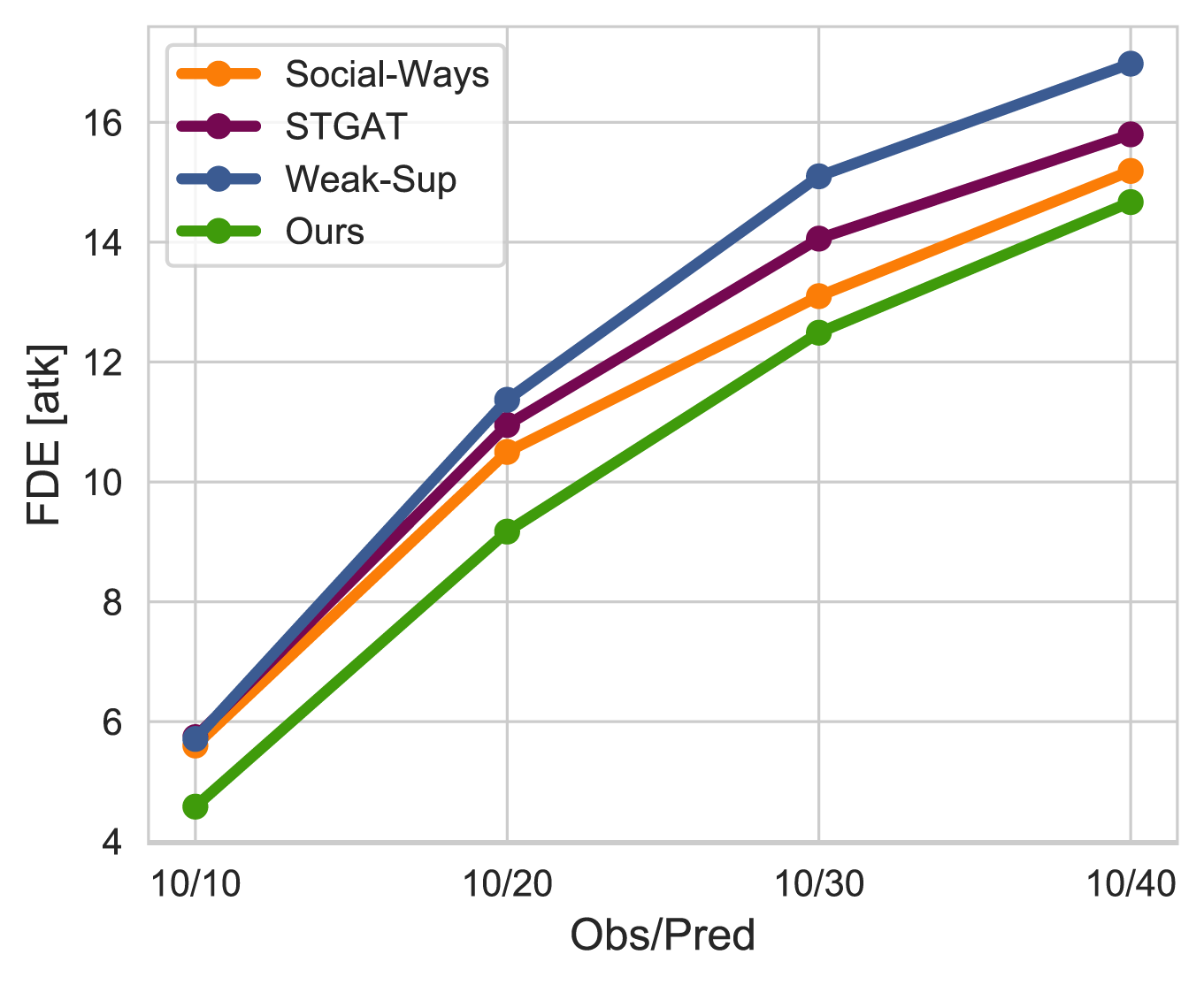}
  \end{minipage}
  \begin{minipage}[b]{0.24\textwidth}
   \includegraphics[width=\textwidth]{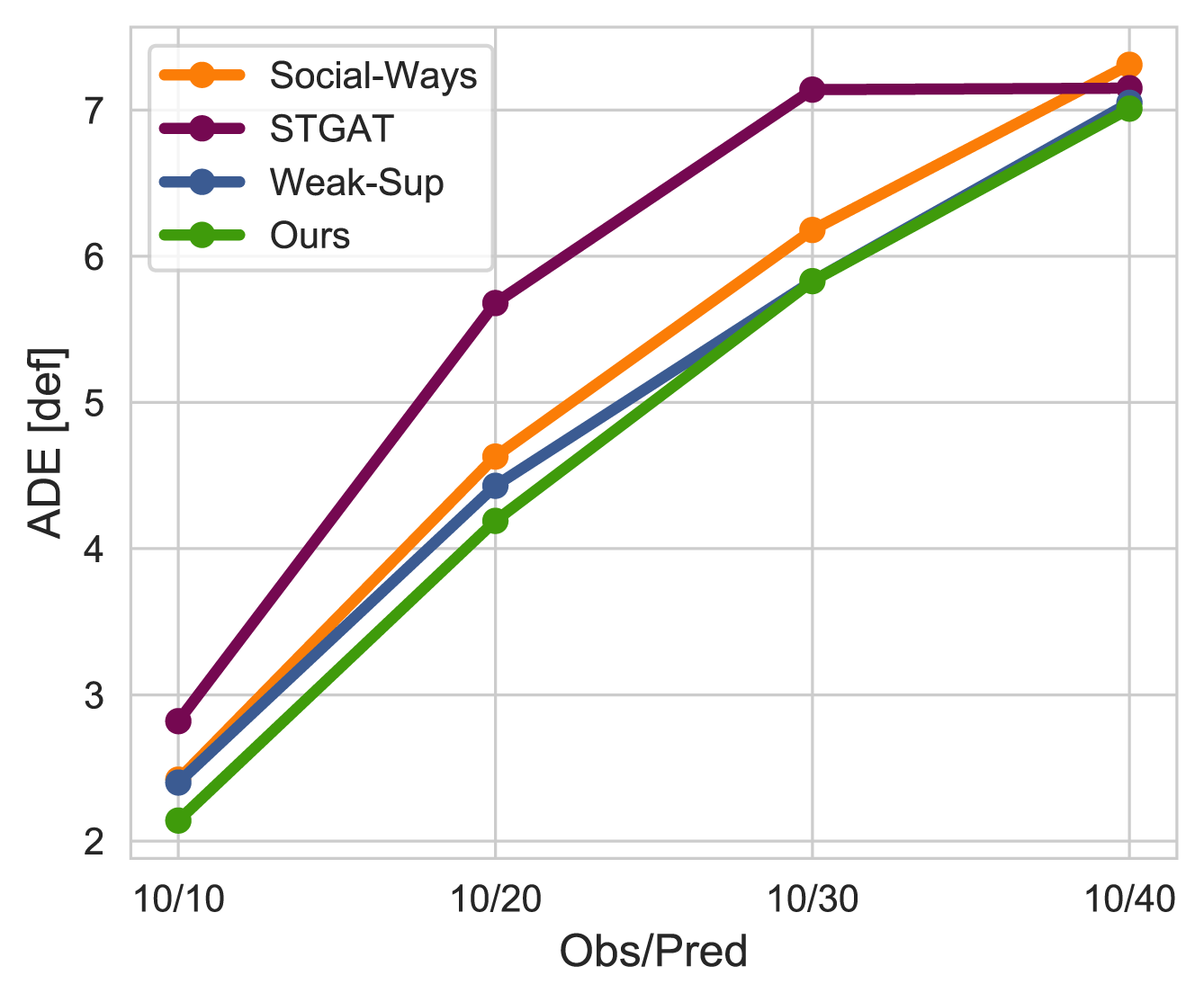}
  \end{minipage}
  \begin{minipage}[b]{0.24\textwidth}
   \includegraphics[width=\textwidth]{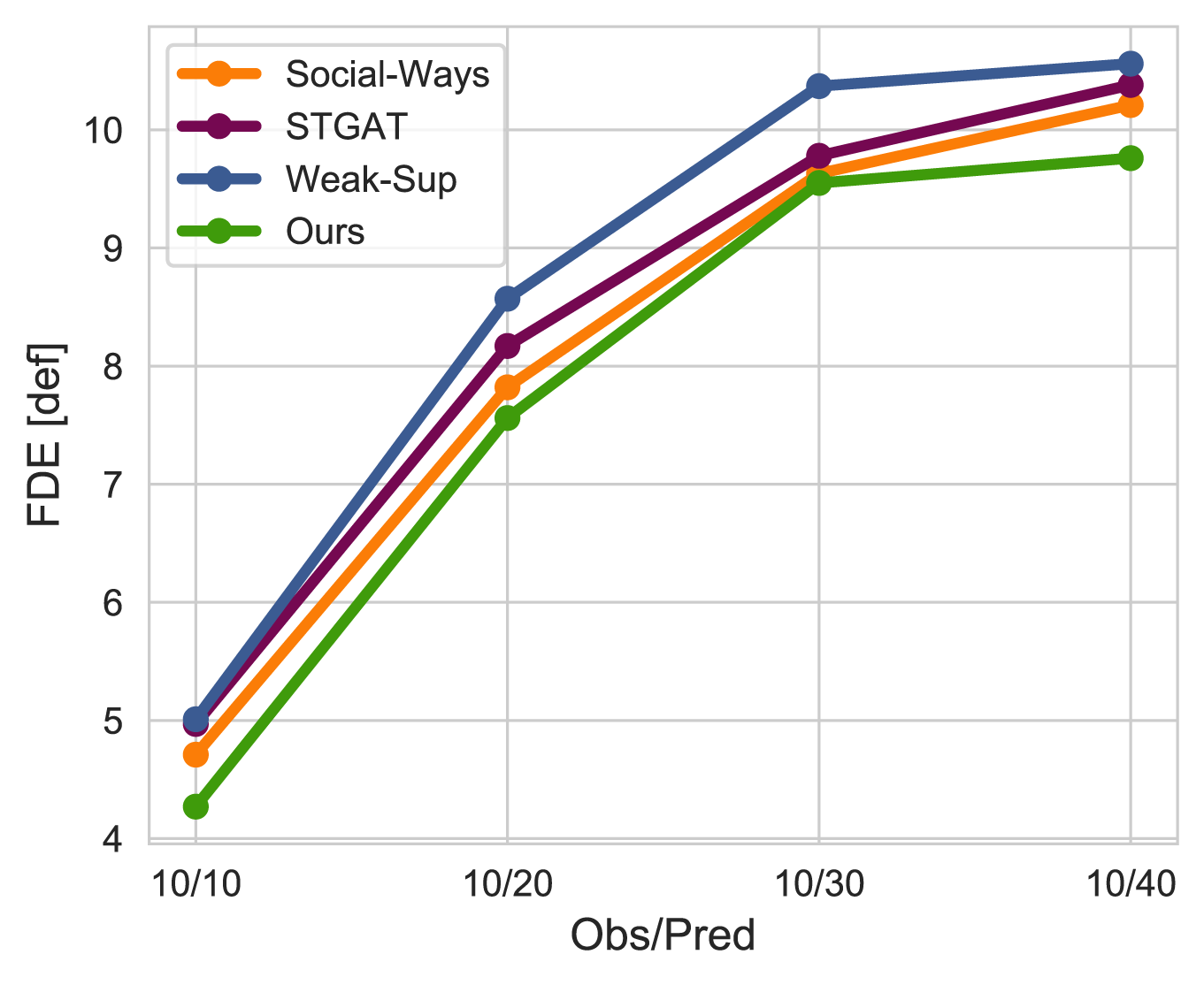}
  \end{minipage}
  \caption{Long-term evaluations: the method is evaluated both in ADE and FDE for increasing prediction lengths, from 10 to 40 time-steps. Attack on the top, defence on the bottom. All the metrics are in feet.}
\label{fig:basketball_long_terms}   
\end{figure}

To evaluate whether our model could show appreciable performance on different prediction horizons, we produced some \textit{long-term evaluations}: since basketball trajectories offered a high number of time-steps with which we could produce various splits, we focused again on sports. For producing such evaluations, we concentrated on different observation-prediction sequences: given 10 time-steps of observation, we evaluated all the methods on increasing prediction splits, from 10 time-steps to 40 time-steps, with steps of 10. As Fig. \ref{fig:basketball_long_terms} shows, our method globally outperforms the competitors in all the different evaluations and in both the metrics. As for the numbers in Table \ref{sportvu_results}, the difference is more pronounced for the attack than for the defence.

We have also run some long-term evaluations considering a longer observation period (Table \ref{ours_vs_cvae_long_term}), mainly to observe how the prediction accuracy changes when the model is allowed to adjust to a greater initial period: we let the model burn-in for 20 initial time-steps and then predict the remaining ones, again with increasing steps of 10. In this setting we are able to compare our model to a further autoencoder architecture, by Felsen et al. \cite{where_will_they_go}, that briefly employs a C-VAE \cite{cvae2} conditioned on players' role. Our model, even without additional information about players' identities, shows better metrics in terms of the average distance from ground-truth positions.

\bigbreak

In urban settings such as the ones in SDD (Table \ref{sdd_results}), the aforementioned distinction between different categories of agents is no longer necessary, because all the pedestrians share the same nature and actively cooperate to not interfere with each other. 
The trajectories are split into segments of 8s: we observe 3.2s of history and predict over a 4.8s future horizon. Operating at 0.4s per time-step results in 8 time-steps of observation and a future prediction span of 12 time-steps; all the metrics are in meters. We can not report results of Weak-Supervision for SDD: since the model adopts a separate VRNN for each agent, its architecture is not suitable for less constrained scenes that exhibit a variable number of agents. For this reason, Weak-Supervision could not be tested outside the basket environment.

The results are in line with the ones presented for basketball. The attentive sharing of agents' goals and the graph distillation step for the hidden states outperforms the competitors in both ADE and FDE: individual goals force agents to pass through specific future areas coherent to their past trajectory, while the two attentive mechanisms help them to keep in consideration both other agents' immediate will and their long-term objectives.

\subsection{Qualitative Results}

\begin{figure}[!t]
  \centering
  \vspace{4 pt}
  \begin{minipage}[b]{0.15\textwidth}
   \includegraphics[width=\textwidth]{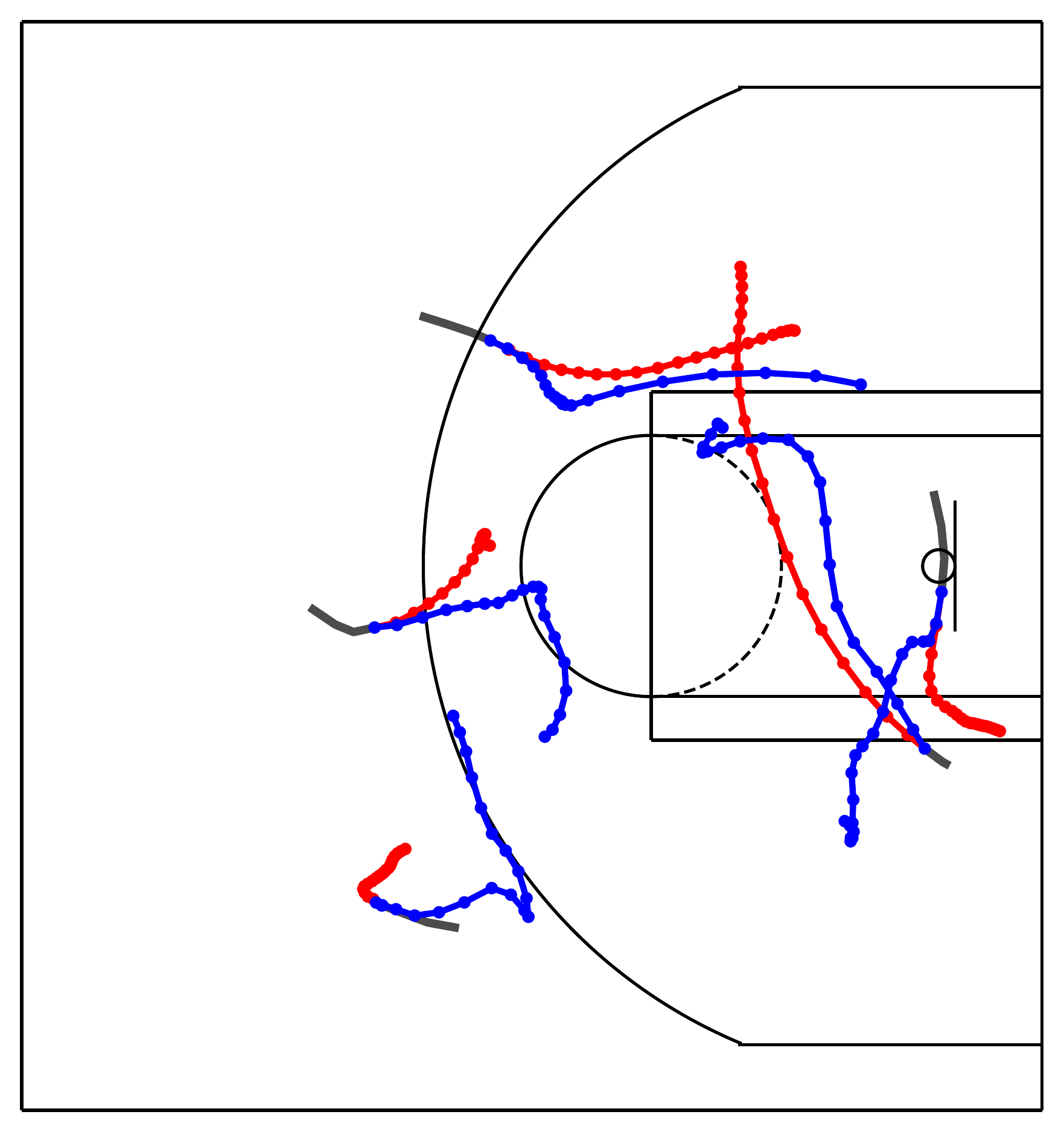}
  \end{minipage}
  \begin{minipage}[b]{0.15\textwidth}
    \includegraphics[width=\textwidth]{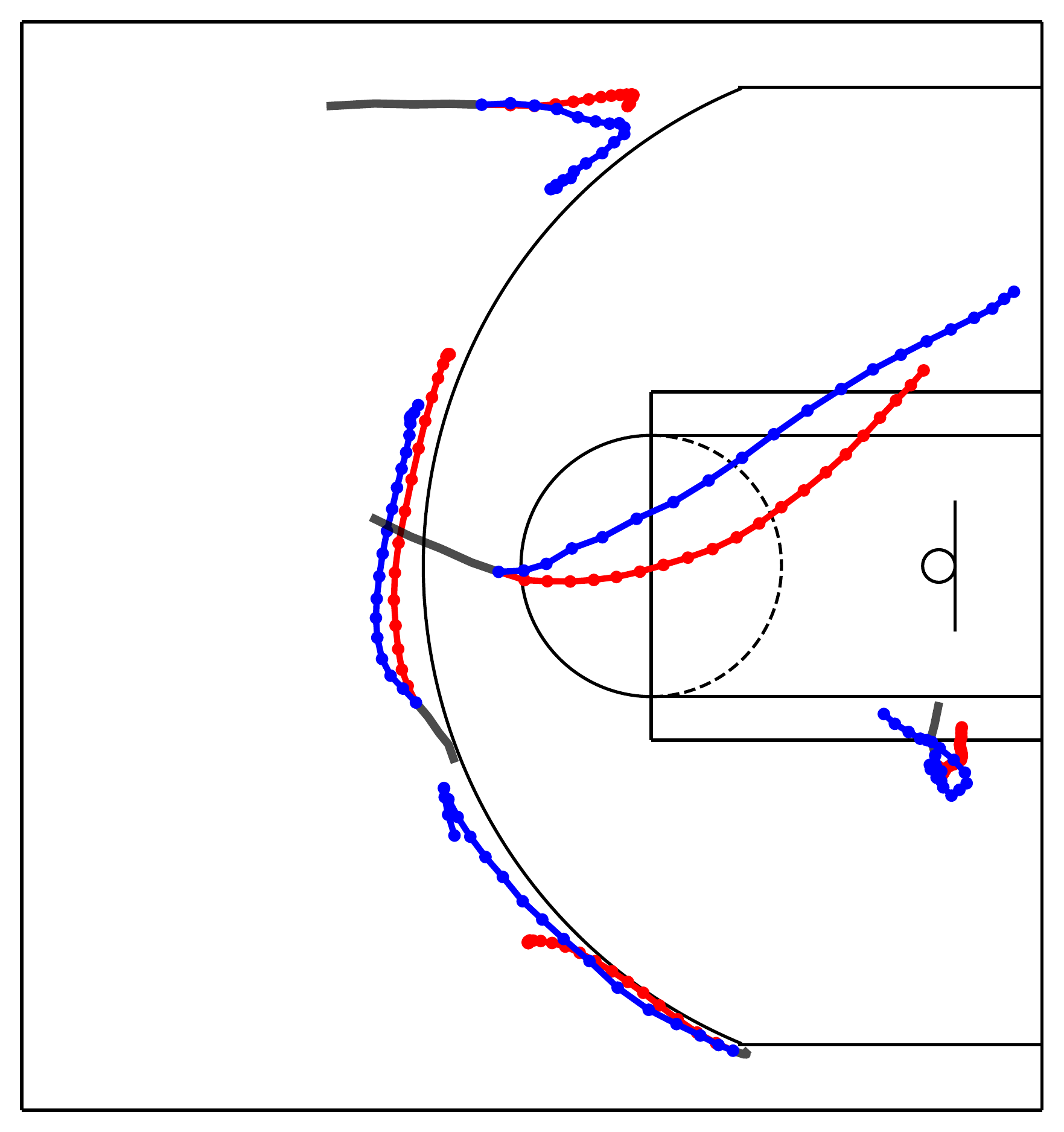}
  \end{minipage}
  \begin{minipage}[b]{0.15\textwidth}
   \includegraphics[width=\textwidth]{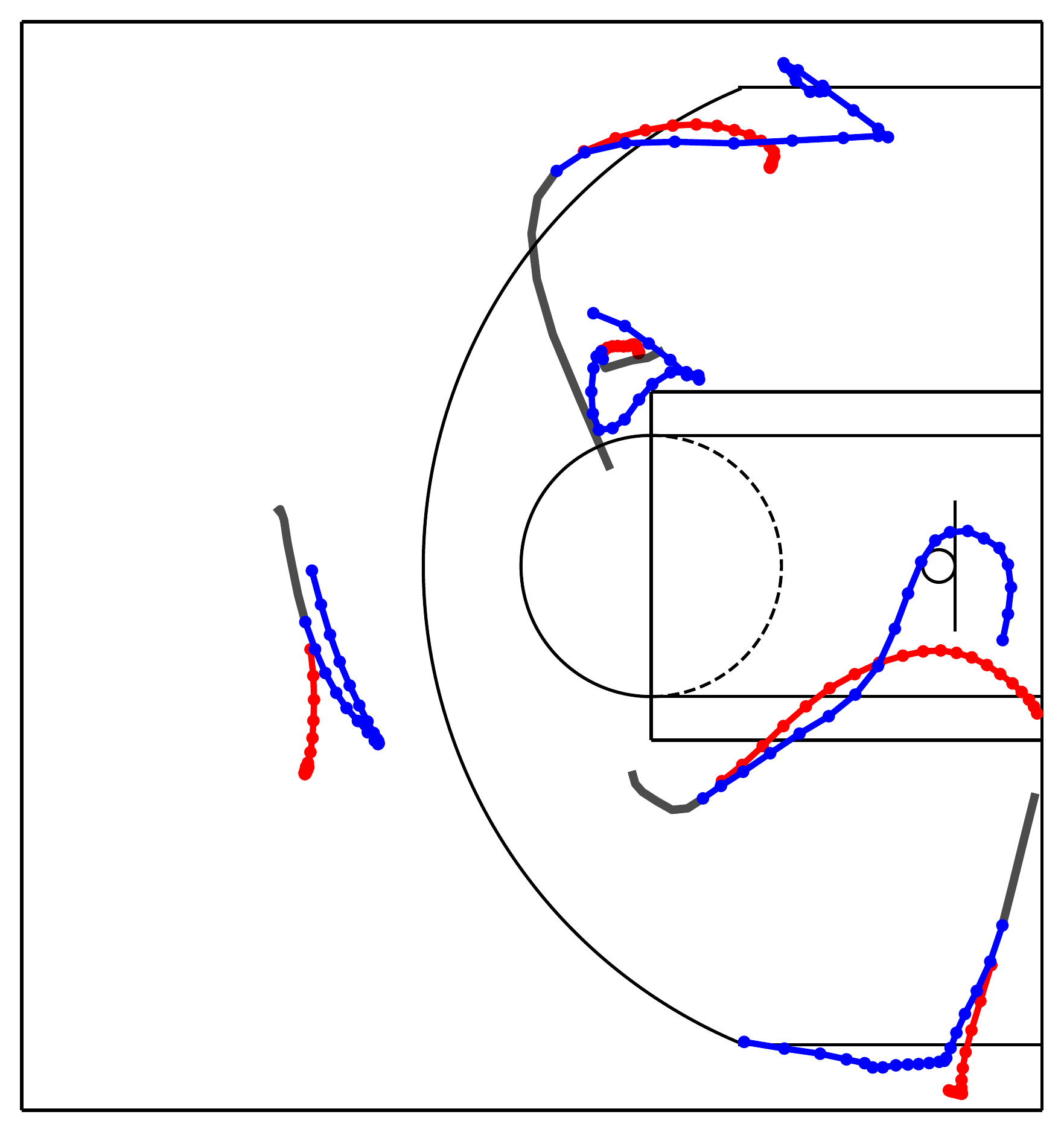}
  \end{minipage}
  \begin{minipage}[b]{0.15\textwidth}
   \includegraphics[width=\textwidth]{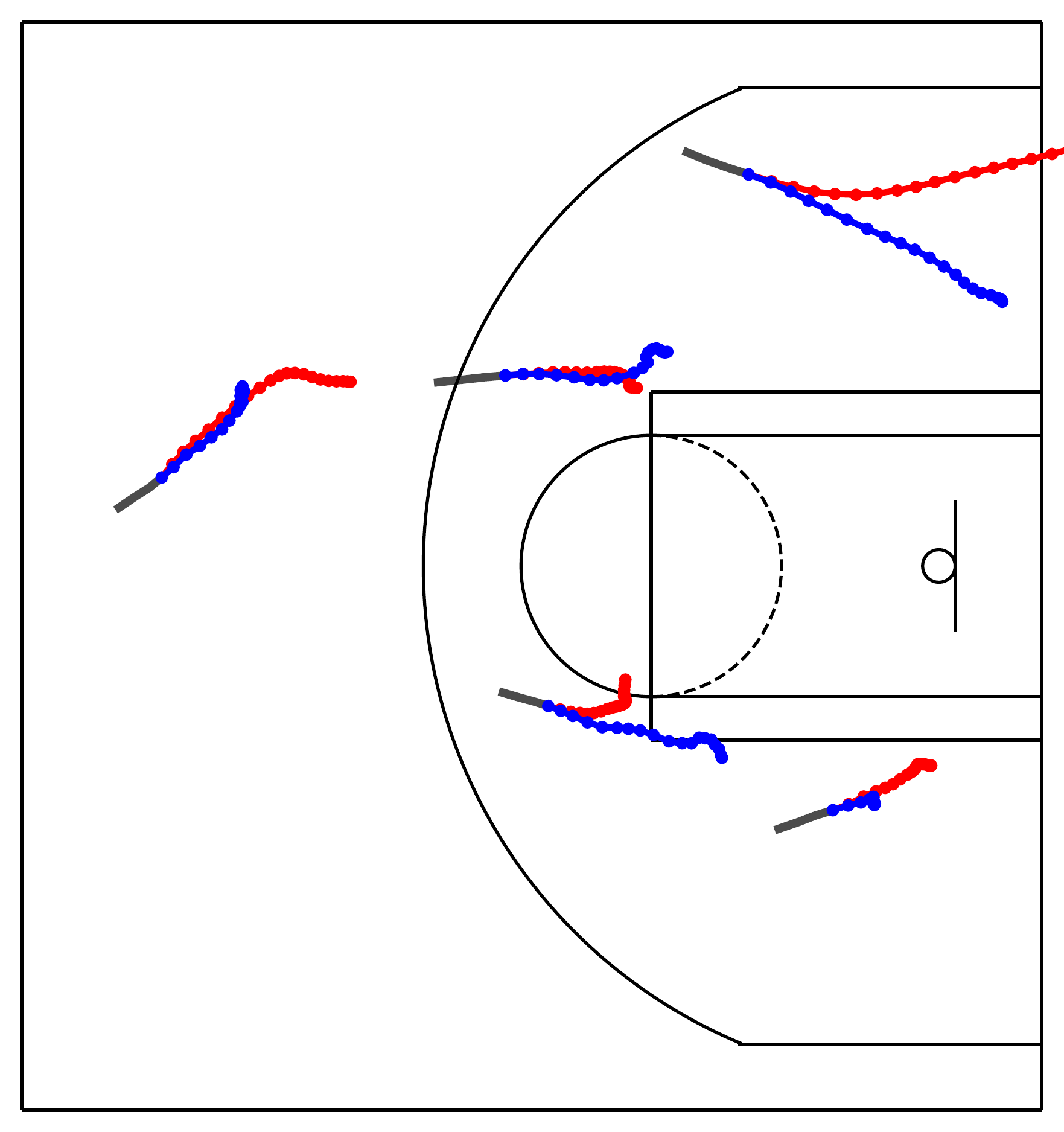}
  \end{minipage}
  \begin{minipage}[b]{0.15\textwidth}
    \includegraphics[width=\textwidth]{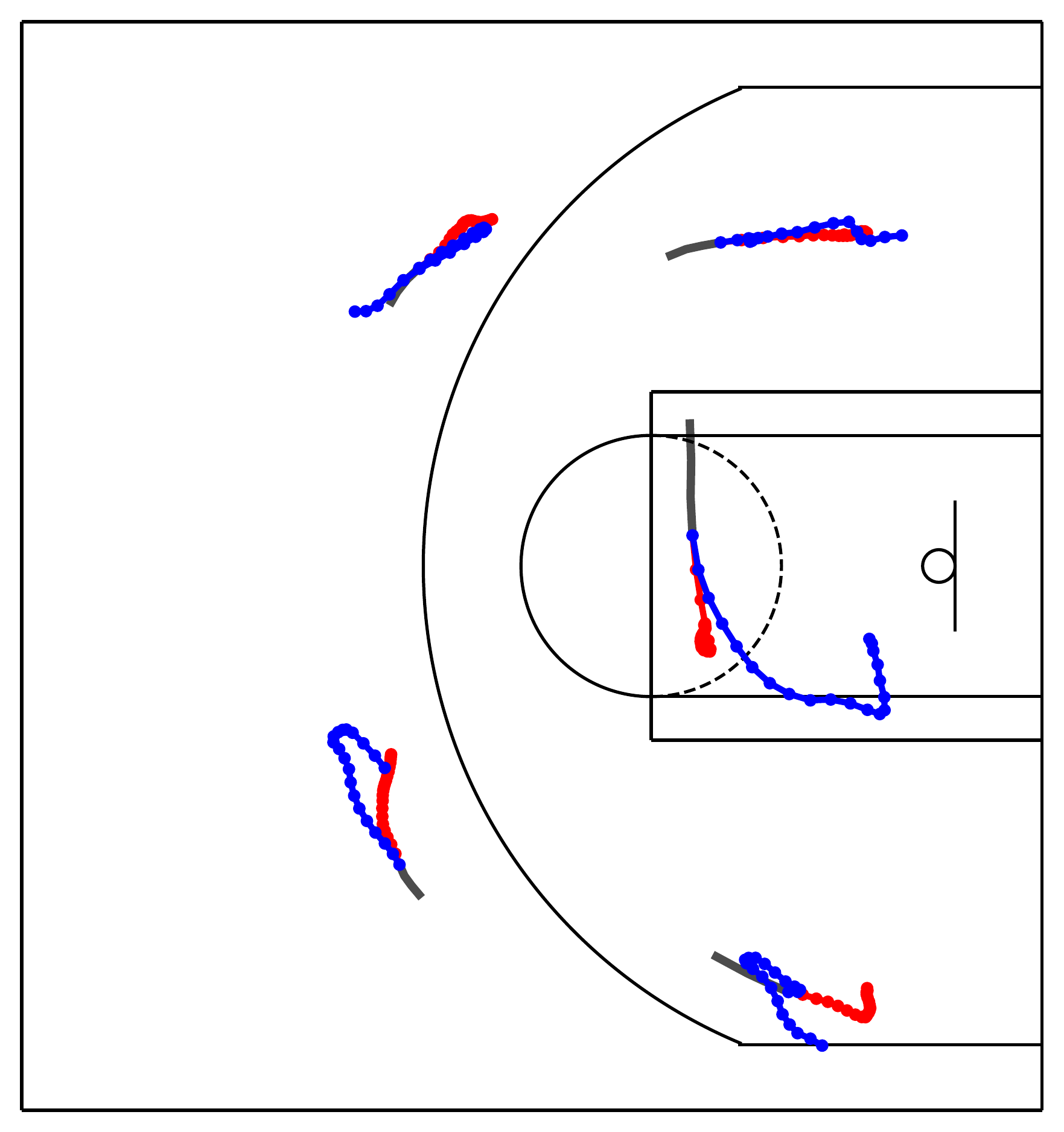}
  \end{minipage}
  \begin{minipage}[b]{0.15\textwidth}
   \includegraphics[width=\textwidth]{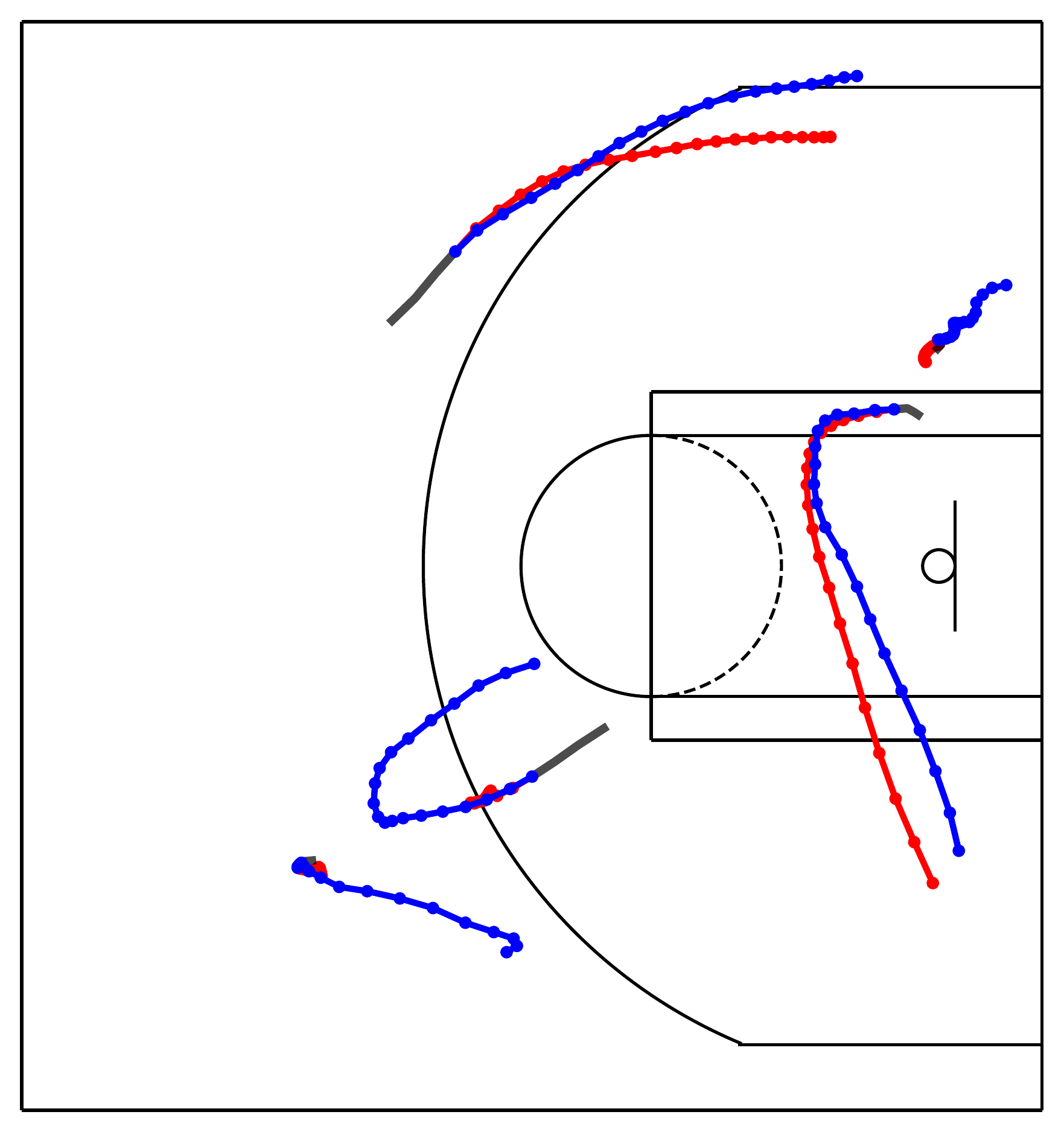}
  \end{minipage}
  \caption{Basketball roll-outs. After an initial observation stage (black), model predictions (red) are evaluated against the ground-truth (blue),  The top roll-outs refer to three different attack plays, while the bottom one represent three different defensive actions.}
  \label{fig:basketball_rollouts}
\end{figure}

\begin{figure}[!t]
  \centering
  \begin{minipage}[b]{0.241\textwidth}
   \includegraphics[width=\textwidth]{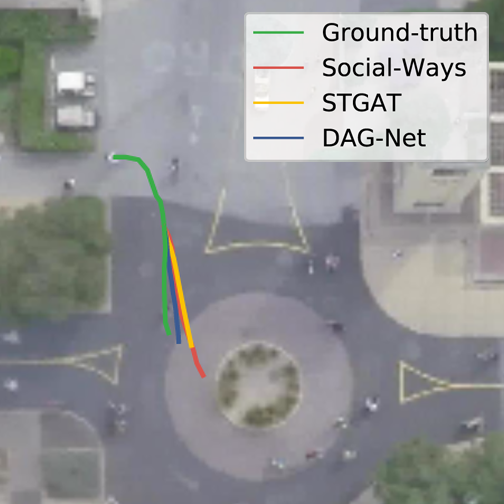}
  \end{minipage}
  \begin{minipage}[b]{0.241\textwidth}
    \includegraphics[width=\textwidth]{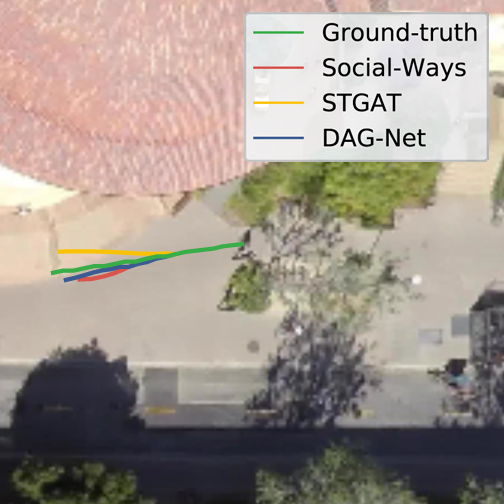}
  \end{minipage}
  \caption{Qualitative samples that compare DAG-Net and state-of-the-art methods on Stanford Drone Dataset.}
  \label{fig:sdd_predictions}
\end{figure}

In competitive settings such as sports, the opposing teams are intrinsically different. The attacking team drives the game trying to score, while the defenders often limit to counter react to its moves. These behaviours deeply affect the resulting trajectories, as can be clearly seen in the roll-outs presented in Fig. \ref{fig:basketball_rollouts}. Attackers trajectories tend to be particularly varied and intricate, often bending and intersecting; on the contrary, defenders tend to move linearly and occasionally deflect to follow an opponent or to close a gap. Despite some sudden changes of direction in the real trajectories, especially for the attackers, our model is able to correctly predict the overall future movement of the players and rather trace the ground-truth. Because of the complexity of such trajectories, the predictions do not always precisely resemble the expected output: nevertheless, even when the predictions fail to follow the real ground-truth trajectories, the model still predicts a likely behaviour coherent with the play development, proving its strength in capturing the multi-modal nature of players' movements.

On the other hand, urban trajectories are more straightforward, because pedestrian obviously tend to move linearly, doing only some occasional deviations to avoid collisions or to turn. Nevertheless, the adoption of agents' goals still gives the model the possibility to produce more likely trajectories. Since agents are constrained to pass through specific portions of the scene coherent with their motion behaviour, predictions can closely resemble real future movements: in both the plot reported in Fig. \ref{fig:sdd_predictions}, DAG-Net is able to keep closer to the ground-truth, while both the competitors tend to predict more linear trajectories and consequently deviate from the expected output.
For the very same reasons, final predictions can also be more precise: having important insights about the regions the agent will occupy in the future can help the model to appropriately predict the overall portion of the scene where the agent will land at the end of his trajectory. DAG-Net predicted final location resembles the agent's real destination, while both the competitors fail to approximately forecast such information.
\section{Ablation Experiments}

\begin{table}[!t]
    \renewcommand{\arraystretch}{1.3}
    \centering
    \caption{STATS SportVU}
    \begin{tabular}{c|c|cc|cc}
        \hline
        \multirow{2}{*}{\textbf{Team}} & \multirow{2}{*}{\textbf{Model}} & \textbf{Agents'} & \textbf{Future} & \multirow{2}{*}{\textbf{ADE}} & \multirow{2}{*}{\textbf{FDE}}\\
         && \textbf{interact.}& \textbf{object.} && \\
        \hline \hline
                & Vanilla VRNN \cite{vrnn} & \ding{55} & \ding{55} & 9.41 & 15.56\\
            ATK & A-VRNN & \ding{51} & \ding{55} & 9.48 & 15.52\\
                & DAG-Net (Our) & \ding{51} & \ding{51} & \textbf{8.98} & \textbf{14.08}\\
        \hline
                & Vanilla VRNN \cite{vrnn} & \ding{55} & \ding{55} & 7.16 & 10.50 \\
            DEF & A-VRNN & \ding{51} & \ding{55} & 7.05 & 10.34\\
                & DAG-Net (Our) & \ding{51} & \ding{51} & \textbf{6.87} & \textbf{9.76}\\
        \hline
    \end{tabular}
    \label{sportvu_results_ablation}
\end{table}

\begin{table}[!t]
    \renewcommand{\arraystretch}{1.3}
    \centering
    \caption{Stanford Drone Dataset}
    \begin{tabular}{c|cc|cc}
        \hline
        \multirow{2}{*}{\textbf{Model}} & \textbf{Agents'} & \textbf{Future} & \multirow{2}{*}{\textbf{ADE}} & \multirow{2}{*}{\textbf{FDE}}\\
         &\textbf{interact.} & \textbf{object.}&& \\
        \hline \hline
             Vanilla VRNN \cite{vrnn} & \ding{55} & \ding{55} & 0.58 & 1.17\\
             A-VRNN & \ding{51} & \ding{55} & 0.56 & 1.14\\ 
             DAG-Net (Our) & \ding{51} & \ding{51} & \textbf{0.53} & \textbf{1.04}\\
        \hline
    \end{tabular}
    \label{sdd_results_ablation}
\end{table}

In this section we present ablation experiments to show the improvements introduced by each component of our model (Table \ref{sportvu_results_ablation} and Table \ref{sdd_results_ablation}). Results present two baselines: the Vanilla VRNN and the Attentive-VRNN (A-VRNN), i.e. a version of our network that presents only the attentive graph for the hidden states refinement. DAG-Net outperforms both baselines on STATS SportVU and SDD. The Vanilla VRNN experiments show that using a stand-alone network without considering interactions between agents does not allow the model to capture the nature of real paths. For this reason, A-VRNN achieves better performance than Vanilla VRNN; still, this version is not able to capture future structured dependencies between agents. The results obtained with DAG-Net highlight the importance of inserting future information into the prediction and combining humans objectives in a structured way.

\section{Conclusions}
We propose a novel architecture called DAG-Net, a double graph-based network that deals with both past interactions and future goals through attentive mechanisms. By facing trajectory prediction as a structured problem, our model overcomes state-of-the-art performances on both STATS SportVU NBA Dataset and Stanford Drone Dataset, proving its strength on team sports and urban contexts. It shows impressive results also on long-term predictions.
Our future work will focus on the application of our model to more general settings, involving time-series forecasting such as finance and health care.

\section*{Acknowledgements}
Funded by the PRIN PREVUE - PRediction of activities and Events by Vision in an Urban Environment” project (CUP E94I19000650001), PRIN National Research Program, MIUR.

\bibliography{./main}

\begin{thebibliography}{10}
\providecommand{\url}[1]{#1}
\csname url@samestyle\endcsname
\providecommand{\newblock}{\relax}
\providecommand{\bibinfo}[2]{#2}
\providecommand{\BIBentrySTDinterwordspacing}{\spaceskip=0pt\relax}
\providecommand{\BIBentryALTinterwordstretchfactor}{4}
\providecommand{\BIBentryALTinterwordspacing}{\spaceskip=\fontdimen2\font plus
\BIBentryALTinterwordstretchfactor\fontdimen3\font minus
  \fontdimen4\font\relax}
\providecommand{\BIBforeignlanguage}[2]{{%
\expandafter\ifx\csname l@#1\endcsname\relax
\typeout{** WARNING: IEEEtran.bst: No hyphenation pattern has been}%
\typeout{** loaded for the language `#1'. Using the pattern for}%
\typeout{** the default language instead.}%
\else
\language=\csname l@#1\endcsname
\fi
#2}}
\providecommand{\BIBdecl}{\relax}
\BIBdecl

\bibitem{forecasting_interactive_dynamics}
W.-C. Ma, D.-A. Huang, N.~Lee, and K.~M. Kitani, ``Forecasting interactive
  dynamics of pedestrians with fictitious play,'' in \emph{Proceedings of the
  IEEE Conference on Computer Vision and Pattern Recognition}, 2017, pp.
  774--782.

\bibitem{multi_agents_fusion}
T.~Zhao, Y.~Xu, M.~Monfort, W.~Choi, C.~Baker, Y.~Zhao, Y.~Wang, and Y.~N. Wu,
  ``Multi-agent tensor fusion for contextual trajectory prediction,'' in
  \emph{Proceedings of the IEEE Conference on Computer Vision and Pattern
  Recognition}, 2019, pp. 12\,126--12\,134.

\bibitem{glmp-realtime}
A.~Bera, S.~Kim, T.~Randhavane, S.~Pratapa, and D.~Manocha, ``Glmp-realtime
  pedestrian path prediction using global and local movement patterns,'' in
  \emph{2016 IEEE International Conference on Robotics and Automation
  (ICRA)}.\hskip 1em plus 0.5em minus 0.4em\relax IEEE, 2016, pp. 5528--5535.

\bibitem{goal_robots}
J.~Mainprice, R.~Hayne, and D.~Berenson, ``Goal set inverse optimal control and
  iterative replanning for predicting human reaching motions in shared
  workspaces,'' \emph{IEEE Transactions on Robotics}, vol.~32, no.~4, pp.
  897--908, 2016.

\bibitem{surveillance_1}
S.~Oh, A.~Hoogs, A.~Perera, N.~Cuntoor, C.-C. Chen, J.~T. Lee, S.~Mukherjee,
  J.~Aggarwal, H.~Lee, L.~Davis \emph{et~al.}, ``A large-scale benchmark
  dataset for event recognition in surveillance video,'' in \emph{CVPR
  2011}.\hskip 1em plus 0.5em minus 0.4em\relax IEEE, 2011, pp. 3153--3160.

\bibitem{abnormal_crowd}
R.~Mehran, A.~Oyama, and M.~Shah, ``Abnormal crowd behavior detection using
  social force model,'' in \emph{2009 IEEE Conference on Computer Vision and
  Pattern Recognition}.\hskip 1em plus 0.5em minus 0.4em\relax IEEE, 2009, pp.
  935--942.

\bibitem{sgan}
A.~Gupta, J.~Johnson, L.~Fei-Fei, S.~Savarese, and A.~Alahi, ``{Social GAN:
  Socially Acceptable Trajectories with Generative Adversarial Networks},'' in
  \emph{IEEE Conference on Computer Vision and Pattern Recognition (CVPR)},
  2018.

\bibitem{social-bigat}
V.~Kosaraju, A.~A. Sadeghian, R.~Mart{\'i}n-Mart{\'i}n, I.~D. Reid, S.~H.
  Rezatofighi, and S.~Savarese, ``Social-bigat: Multimodal trajectory
  forecasting using bicycle-gan and graph attention networks,'' in \emph{IEEE
  Conference on Computer Vision and Pattern Recognition (CVPR)}, 2019.

\bibitem{slstm}
A.~{Alahi}, K.~{Goel}, V.~{Ramanathan}, A.~{Robicquet}, L.~{Fei-Fei}, and
  S.~{Savarese}, ``{Social LSTM: Human Trajectory Prediction in Crowded
  Spaces},'' in \emph{2016 IEEE Conference on Computer Vision and Pattern
  Recognition (CVPR)}, June 2016, pp. 961--971.

\bibitem{vae}
D.~P. Kingma and M.~Welling, ``Auto-encoding variational bayes,'' \emph{CoRR},
  vol. abs/1312.6114, 2013.

\bibitem{socialforce}
D.~Helbing and P.~Molnar, ``Social force model for pedestrian dynamics,''
  \emph{Physical review E}, vol.~51, no.~5, p. 4282, 1995.

\bibitem{discrete_choice}
G.~Antonini, M.~Bierlaire, and M.~Weber, ``{Discrete Choice Models for
  Pedestrian Walking Behavior},'' \emph{Transportation Research Part B:
  Methodological}, vol.~40, pp. 667--687, 09 2006.

\bibitem{continuum_dynamics}
A.~Treuille, S.~Cooper, and Z.~Popovic, ``Continuum crowds,'' \emph{ACM Trans.
  Graph.}, vol.~25, pp. 1160--1168, 07 2006.

\bibitem{gaussian_process}
J.~Wang, A.~Hertzmann, and D.~J. Fleet, ``{Gaussian Process Dynamical
  Models},'' in \emph{Advances in Neural Information Processing Systems 18},
  Y.~Weiss, B.~Sch\"{o}lkopf, and J.~C. Platt, Eds.\hskip 1em plus 0.5em minus
  0.4em\relax MIT Press, 2006, pp. 1441--1448.

\bibitem{soft+}
T.~Fernando, S.~Denman, S.~Sridharan, and C.~Fookes, ``Soft+ hardwired
  attention: An lstm framework for human trajectory prediction and abnormal
  event detection,'' \emph{Neural networks}, vol. 108, pp. 466--478, 2018.

\bibitem{desire}
N.~Lee, W.~Choi, P.~Vernaza, C.~B. Choy, P.~H.~S. Torr, and M.~K. Chandraker,
  ``Desire: Distant future prediction in dynamic scenes with interacting
  agents,'' \emph{2017 IEEE Conference on Computer Vision and Pattern
  Recognition (CVPR)}, pp. 2165--2174, 2017.

\bibitem{socialways}
A.~Javad, H.~Jean-Bernard, and P.~Julien, ``Social ways: Learning multi-modal
  distributions of pedestrian trajectories with gans,'' in \emph{Proceedings of
  the IEEE Conference on Computer Vision and Pattern Recognition Workshops
  (CVPRW)}, 2019, pp. 0--0.

\bibitem{stgat}
H.~Yingfan, B.~Huikun, L.~Zhaoxin, M.~Tianlu, and W.~Zhaoqi, ``{STGAT: Modeling
  Spatial-Temporal Interactions for Human Trajectory Prediction},'' in
  \emph{The IEEE International Conference on Computer Vision (ICCV)}, October
  2019.

\bibitem{gat}
P.~Veli{\v{c}}kovi{\'{c}}, G.~Cucurull, A.~Casanova, A.~Romero, P.~Li{\`{o}},
  and Y.~Bengio, ``{Graph Attention Networks},'' \emph{International Conference
  on Learning Representations (ICLR)}, 2018.

\bibitem{attention_is_all_you_need}
A.~Vaswani, N.~Shazeer, N.~Parmar, J.~Uszkoreit, L.~Jones, A.~N. Gomez, L.~u.
  Kaiser, and I.~Polosukhin, ``Attention is all you need,'' in \emph{Advances
  in Neural Information Processing Systems 30}, I.~Guyon, U.~V. Luxburg,
  S.~Bengio, H.~Wallach, R.~Fergus, S.~Vishwanathan, and R.~Garnett, Eds.\hskip
  1em plus 0.5em minus 0.4em\relax Curran Associates, Inc., 2017, pp.
  5998--6008.

\bibitem{weeksup}
E.~Zhan, S.~Zheng, Y.~Yue, L.~Sha, and P.~Lucey, ``Generating multi-agent
  trajectories using programmatic weak supervision,'' in \emph{International
  Conference on Learning Representations (ICLR)}, 2019.

\bibitem{deep_gen_models}
D.~J. Rezende, S.~Mohamed, and D.~Wierstra, ``{Stochastic Backpropagation and
  Approximate Inference in Deep Generative Models},'' in \emph{{Proceedings of
  the 31st International Conference on Machine Learning, Cycle 2}}, ser. {JMLR
  Proceedings}, vol.~32.\hskip 1em plus 0.5em minus 0.4em\relax {JMLR.org},
  2014, pp. 1278--1286.

\bibitem{vrnn}
J.~Chung, K.~Kastner, L.~Dinh, K.~Goel, A.~Courville, and Y.~Bengio, ``{A
  Recurrent Latent Variable Model for Sequential Data},'' in \emph{Proceedings
  of the 28th International Conference on Neural Information Processing Systems
  - Volume 2}, ser. NIPS'15.\hskip 1em plus 0.5em minus 0.4em\relax Cambridge,
  MA, USA: MIT Press, 2015, pp. 2980--2988.

\bibitem{cvae}
D.~P. Kingma, D.~J. Rezende, S.~Mohamed, and M.~Welling, ``Semi-supervised
  learning with deep generative models,'' in \emph{Proceedings of the 27th
  International Conference on Neural Information Processing Systems - Volume
  2}, ser. NIPS'14.\hskip 1em plus 0.5em minus 0.4em\relax Cambridge, MA, USA:
  MIT Press, 2014, pp. 3581--3589.

\bibitem{cvae2}
K.~Sohn, X.~Yan, and H.~Lee, ``Learning structured output representation using
  deep conditional generative models,'' in \emph{Proceedings of the 28th
  International Conference on Neural Information Processing Systems - Volume
  2}, ser. NIPS'15.\hskip 1em plus 0.5em minus 0.4em\relax Cambridge, MA, USA:
  MIT Press, 2015, pp. 3483--3491.

\bibitem{conditionalflow}
A.~Bhattacharyya, M.~Hanselmann, M.~Fritz, B.~Schiele, and C.-N. Straehle,
  ``Conditional flow variational autoencoders for structured sequence
  prediction,'' \emph{arXiv}, vol. abs/1908.09008, 2019.

\bibitem{where_will_they_go}
P.~Felsen, P.~Lucey, and S.~Ganguly, ``Where will they go? predicting
  fine-grained adversarial multi-agent motion using conditional variational
  autoencoders,'' in \emph{ECCV}, 2018.

\bibitem{Jiachen_IROS19}
L.~Jiachen, M.~Hengbo, and T.~Masayoshi, ``Conditional generative neural system
  for probabilistic trajectory prediction,'' in \emph{in 2019 IEEE/RSJ
  International Conference on Intelligent Robots and Systems (IROS)}.\hskip 1em
  plus 0.5em minus 0.4em\relax IEEE, 2019.

\bibitem{sdd}
A.~Robicquet, A.~Alahi, A.~Sadeghian, B.~Anenberg, J.~Doherty, E.~Wu, and
  S.~Savarese, ``{Forecasting Social Navigation in Crowded Complex Scenes},''
  \emph{CoRR}, vol. abs/1601.00998, 2016.

\bibitem{trajnet2018}
A.~Sadeghian, V.~Kosaraju, A.~Gupta, S.~Savarese, and A.~Alahi, ``Trajnet:
  Towards a benchmark for human trajectory prediction,'' \emph{arXiv preprint},
  2018.

\bibitem{stats_sportvu}
``{SportVU - STATS Perform},''
  \url{https://www.statsperform.com/team-performance/basketball/optical-tracking/}.

\end{thebibliography}
\bibliographystyle{IEEEtran}

\end{document}